\documentclass{article}

\usepackage[preprint]{neurips_2021}

\usepackage[utf8]{inputenc} 
\usepackage[T1]{fontenc}    
\usepackage{hyperref}       
\usepackage{url}            
\usepackage{booktabs}       
\usepackage{amsfonts}       
\usepackage{nicefrac}       
\usepackage{microtype}      

\usepackage{amsmath, mathtools}      
\usepackage{stackengine}
\usepackage{bm} 
\def\delequal{\mathrel{\ensurestackMath{\stackon[1pt]{=}{\scriptstyle\Delta}}}}

\usepackage[dvipsnames]{xcolor} 
\usepackage[group-separator={,}]{siunitx}

\usepackage{caption}
\usepackage{subcaption}
\usepackage{graphicx}

\usepackage{adjustbox}
\usepackage{multirow}

\title{GroupBERT: Enhanced Transformer Architecture with Efficient Grouped Structures }

\author{
   Ivan Chelombiev \thanks{Equal contribution.}\\
   Graphcore Research \\
   \texttt{ivanc@graphcore.ai} \\
   \And
    Daniel Justus \footnotemark[1] \\
   Graphcore Research \\
   \texttt{danielj@graphcore.ai} \\
   \And
   Douglas Orr \footnotemark[1]\\
   Graphcore Research \\
   \texttt{douglaso@graphcore.ai} \\
    \And
   \hspace{1mm} Anastasia Dietrich \\
   \hspace{1mm} Graphcore Research \\
   \hspace{1mm} \texttt{anastasiad@graphcore.ai} \\
    \And
   \hspace{-10mm} Frithjof Gressmann \\
   \hspace{-10mm} Graphcore Research \\
   \hspace{-10mm} \texttt{frithjof@graphcore.ai} \\
    \And
   \hspace{-14mm} Alexandros Koliousis\thanks{Work done while at Graphcore.} \\
   \hspace{-14mm} New College of the Humanities \\
   \hspace{-14mm} \texttt{alexandros.koliousis@nchlondon.ac.uk} \\
    \And
   \hspace{-14mm}Carlo Luschi \\
   \hspace{-14mm}Graphcore Research \\
   \hspace{-14mm}\texttt{carlo@graphcore.ai} \\
}

\begin{document}

\maketitle

\begin{abstract}
Attention based language models have become a critical component in state-of-the-art natural language processing systems. However, these models have significant computational requirements, due to long training times, dense operations and large parameter count. In this work we demonstrate a set of modifications to the structure of a Transformer layer, producing a more efficient architecture. First, we add a convolutional module to complement the self-attention module, decoupling the learning of local and global interactions. Secondly, we rely on grouped transformations to reduce the computational cost of dense feed-forward layers and convolutions, while preserving the expressivity of the model.  We apply the resulting architecture to language representation learning and demonstrate its superior performance compared to BERT models of different scales. We further highlight its improved efficiency, both in terms of floating-point operations (FLOPs) and time-to-train.

\end{abstract}

\section{Introduction}
Deep neural networks have emerged as the leading solution to enabling end-to-end language processing \citep{lstm, nmt,  gru}. Recently, the Transformer model based on the self-attention mechanism~\citep{vaswani} has become the most promising architecture for language applications \citep{bert, gpt2, gpt3}. Attention based models are also increasingly showing promising results for established applications in domains different from natural language processing~\citep{16by16}.

Complementary to the Transformer's improved ability to model long-range dependencies in sequences is its superior potential to scale to larger sizes \citep{kaplan} and its suitability for execution on existing accelerators. This makes these models favoured over traditional recurrent language models. Given the increased computational demand of these models, there is a growing and pressing interest to develop more efficient architectures \citep{climate_paper}. Some previous proposals were able to reduce the computational burden of the Transformer with improved task performance, but often with a corresponding slower execution, as will be discussed further in Section~\ref{sec:litreview}. While these models might not have passed the Hardware Lottery filter \citep{hardware_lottery}, in this work we leverage Graphcore's Intelligence Processing Unit (IPU)~\citep{ipu}, which has allowed us to examine a wide variety of techniques, centered around improving computational efficiency.

We demonstrate a set of modifications to the the structure of the Transformer layer that improve FLOP utilization by the encoder stack. The proposed \emph{GroupBERT} model relies on grouped matrix multiplications and convolutions, and delivers a more efficient version of the BERT architecture, superior in both task performance and computation efficiency. These efficient building blocks have a reduced computational load for a given memory access~\citep{Masters21}. This property would make them undesirable for traditional accelerators, which rely on large dense computations and reduced  memory access. However, the IPU hardware architecture uses on-chip SRAM for model execution, which opens up the possibility of using more efficient computation blocks that would be discarded by users of hardware with smaller memory bandwidth.


We achieve a performance boost by extending each Transformer layer to contain four modules: one multi-head attention (MHA), one grouped convolution module, and two grouped feed-forward modules (GFFN). The MHA and grouped convolution modules process token information along the sequence dimension, and each is followed by the general computation GFFN module. While there are twice as many modules in the proposed GroupBERT layer, the overall increase in computation is  modest as we utilize sparse grouped operations, for a total FLOP increase of about $60\%$. 

 Not only does GroupBERT deliver better performance per FLOP, but it is also executed faster as measured in total time-to-train. By employing both attention and convolution, the model has components dedicated to both short and long-range interactions, making a more efficient use of the more expensive attention mechanism. We also utilize the parameters of GroupBERT more efficiently during training, by discarding dropout for pre-training on a large corpus of text and by improving stability to use higher learning rates. With all these innovations, GroupBERT Base is only slightly larger than BERT Base, yet it achieves better validation MLM loss than BERT Large using less than half of its FLOPs.

\section{Related Work}\label{sec:litreview}
Grouped transformations have been prevalent in Convolutional Neural Networks (CNN), initially used for their parameter efficiency, starting with Alexnet \citep{alexnet}. They have been further popularized by the ResNeXt architecture~\citep{resnext}. More recent CNN architectures implement depthwise separable convolutions, which are a special case of grouped convolutions with group size equal to one \citep{mobilenet, efficientnet}. In addition to being beneficial for resource utilization, the use of grouped transformation was found to be particularly well suited to convolutional neural network architectures \citep{Ioannou17}.

In language modelling, modern attention based architectures utilize most of their non-embedding parameters in dense operations, thus driving up the computational requirements of these models. Some studies have already considered the use of grouped operations in Transformer-XL~\citep{transformerxl}: DeLighT \citep{delight} and DeFINE \citep{define} use cascades of grouped transformations in the encoder and the embedding, respectively, to lower the computational load of the Transformer. While sophisticated, these methods do not provide a compelling acceleration for the model execution. On the contrary, SqueezeBERT \citep{squeezebert} replaces almost all dense multiplications with grouped transformations in a Transformer layer. This approach shows a significant speedup, but loses too much task performance if pre-trained without distillation from a dense parent model~\citep{distillation}. In this work, we present a more balanced implementation of grouped transformations, delivering both performance and speed.

The Transformer model was initially introduced as a parallelizable solution for supervised sequence-to-sequence learning in the field of natural language processing. These models have become ubiquitous with the use of self-supervised approaches \citep{bert, gpt2}. While attention based models have superior performance due to their ability to model long-range interactions, this characteristic is also what makes the attention mechanism costly. Many studies have succesfully managed to  make the attention mechanism more efficient \citep{efficientreview, transformer_review}. However, its general use across the model can be redundant, as some attention heads in practice reduce to convolutions to model local interactions~\citep{Cordonnier2020On} and duplicate other attention heads, making them redundant~\citep{headprune}. Using convolutions directly inside a Transformer has been shown to make attention more focused on long-range interactions, thus enabling more of its capacity to be used for that specific purpose \citep{lightweightconv, convbert, Wu2020Lite}. While the incorporation of convolutions in Transformers was previously investigated only for smaller models with no more than 100M parameters, we look at scaling to a much wider range of model sizes, from 30M to 500M parameters.

\section{Architecture}
\begin{figure}[t]
\centering
\begin{subfigure}[t]{0.63\linewidth}
    \includegraphics[width=\linewidth]{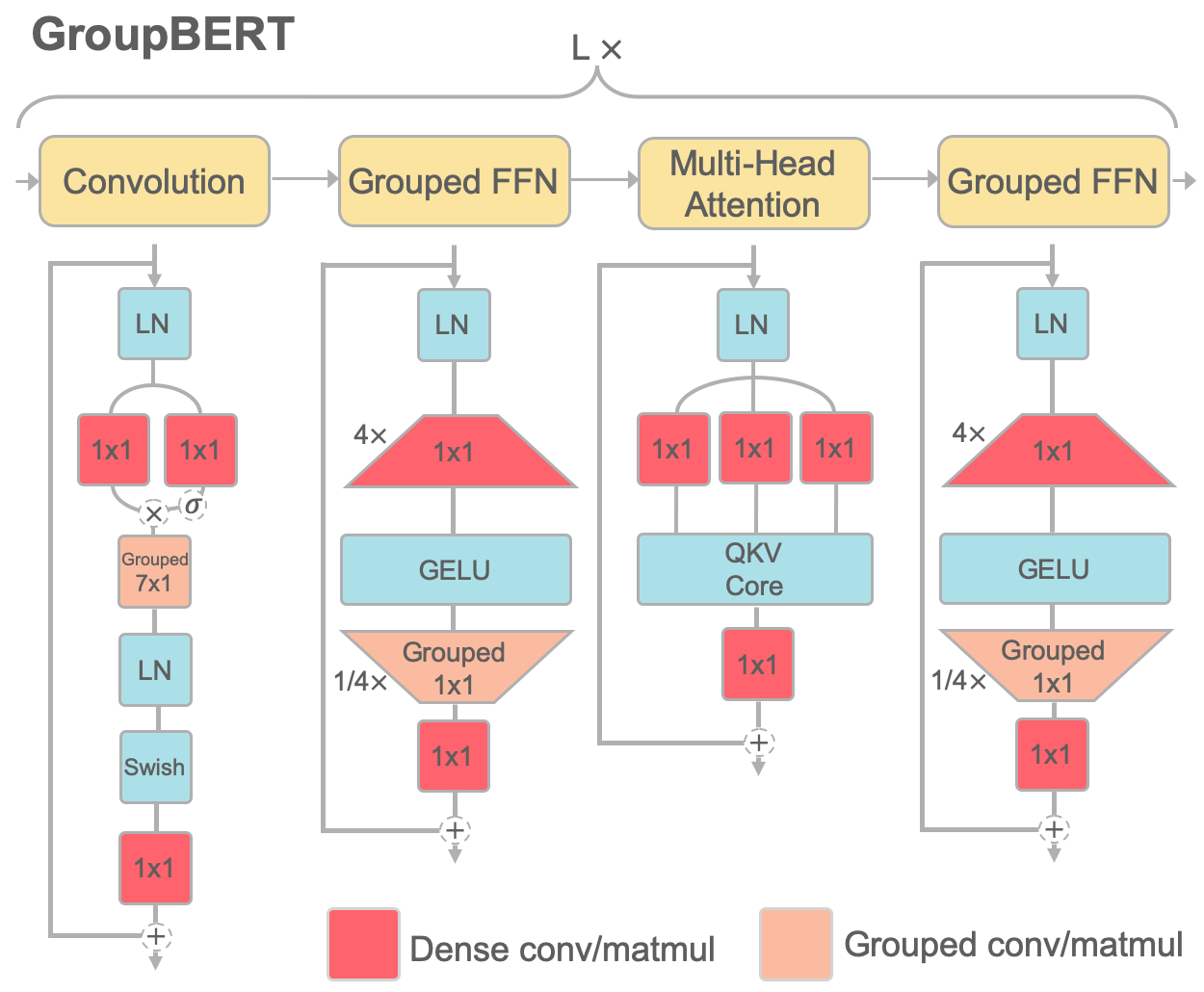}%
    \label{fig:groupbert}%
\end{subfigure}
\quad
\begin{subfigure}[t]{0.33\linewidth}
    \vspace*{-7.29cm}
    \includegraphics[width=\linewidth,height=7.32cm]{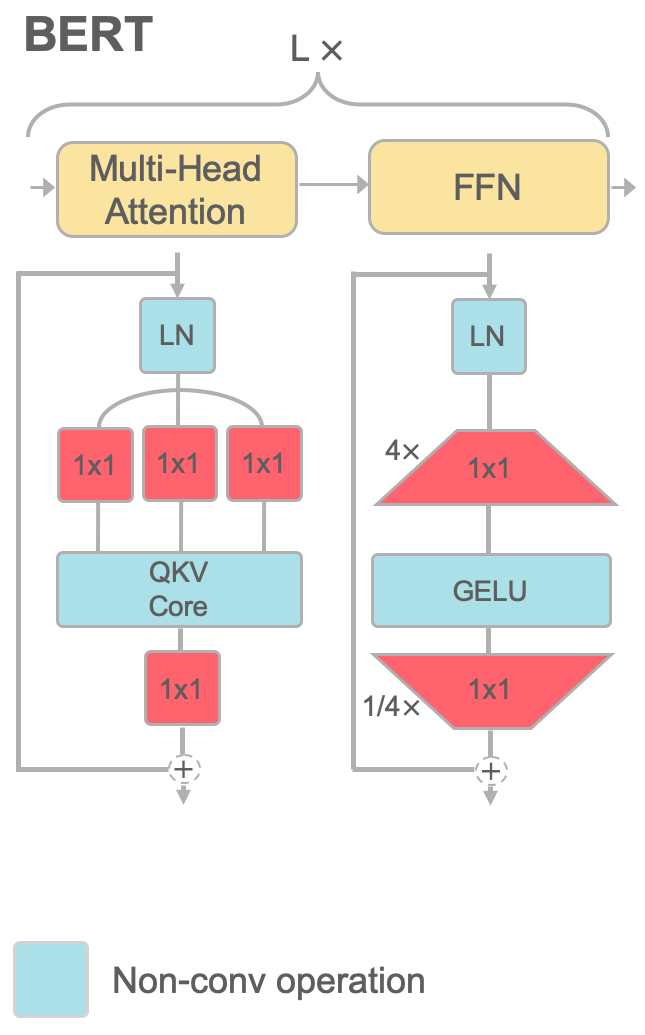}%
    \label{fig:bert}%
\end{subfigure}
\caption{Schematic representations outlining the difference between GroupBERT and BERT structures. A single GroupBERT layer has double the number of modules.}
\label{fig:groupbert_vs_bert}
\end{figure}

In this work, we propose an efficient modification of the Transformer layer called GroupBERT. The original Transformer layer consists of two modules: multi-head attention (MHA) and feed-forward network (FFN). Each of these modules also includes dropout, a shortcut connection, and layer normalization \citep{dropout, resnet, layernorm}. GroupBERT includes four modules in every layer, as illustrated in Figure \ref{fig:groupbert_vs_bert}. We add a convolution module in sequence with the MHA to efficiently model local interactions between tokens and to allow the attention mechanism to focus on long-range interactions. We then complement every sequence processing block with a dedicated fully-connected module. For better efficiency, we introduce grouped projections to the FLOPs intensive FFN module, making the layer structure more FLOP efficient.

\subsection{Convolution Block}
Sequential locality plays an important role for contextualizing tokens in language models. At the same time, long-range interactions have proven to be vital for state-of-the-art performance. Transformers inherently support long-range content-based interactions via self-attention and usually incorporate a form of positional encoding, allowing attention to also capture position-based interactions \citep{transformerxl}. Although this gives self-attention strong representational power, a convolution is a more efficient implementation of strictly local, position-based fusion. For this reason we adopt a dedicated convolutional module to improve overall efficiency.

The design of our convolution module is similar to \citet{Gulati20}, in which convolutions were introduced into a speech recognition Transformer. We apply a gate consisting of a pointwise convolution followed by a Gated Linear Unit (GLU) that has been beneficial in language applications \citep{glu, Wu19, Wu2020Lite}. Unlike \citet{Gulati20}, we use grouped convolutions in place of depthwise convolutions to add representational capacity. We find that the best trade-off between task performance and computational cost is achieved by using a grouped convolution with group size $16$ and kernel size $7$, computed over the sequence dimension. The module also includes an additional layer normalization and a Swish activation \citep{swish}.

\begin{figure}[t]
\centering
\begin{subfigure}[t]{0.27\linewidth}
    \includegraphics[width=\linewidth]{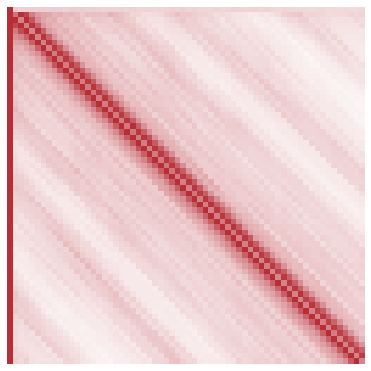}%
    \caption{}
    \label{fig:just_attention}%
\end{subfigure}
\quad
\begin{subfigure}[t]{0.33\linewidth}
    \includegraphics[width=\linewidth]{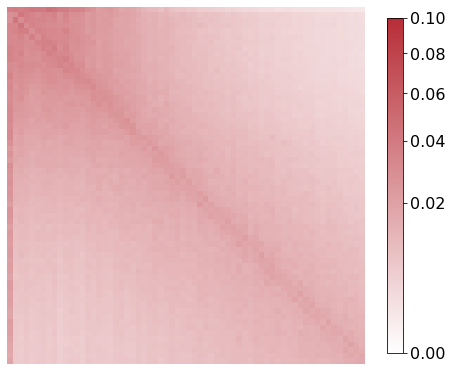}%
    \caption{}
    \label{fig:attention_with_conv}%
\end{subfigure}
\caption{
Attention maps of a single head from the first MHA module of BERT Base (a) and Base~+~Conv (b) (Section \ref{sec:ablation}), averaged over $10^3$ validation sequences to show the sequential locality preference of a typical attention head (see Appendix~\ref{app:attention_maps} for all heads). Each row corresponds to a query token position, each column to a key position. Adding the convolution module encourages attention layers to learn long-range content-based interactions over short-range position-based interactions.}
\label{fig:attention_maps}%
\end{figure}

With this module included, fewer attention heads show a strong locality preference since such interactions are readily captured by convolutions. This effect is visible in the attention maps of Figure~\ref{fig:attention_maps}, showing weaker locality in the model that includes convolutions. To measure this effect quantitatively, we calculate the entropy across target positions for each head and source position. We then average, and normalize by the maximum possible value (see Appendix~\ref{app:attention_maps}). For this measure, zero means that every head attends to a single position exclusively, while one means that every head is position agnostic, although there could still be a joint position and content term. BERT Base has an average entropy ratio of $0.75$ and BERT Base~+~Conv has $0.92$, indicating a shift of positional fusion work from attention to convolution.

\subsection{Grouped Feed-Forward Modules}
The FFN module plays a crucial part in the unparalleled task performance of Transformers~\citep{not_attention, fnet}. Although it is an essential complement to sequence processing modules it introduces a computational burden, since two thirds of the FLOPs are concentrated in the FFN module. To make it more lightweight we utilize structured sparsity in a form of sparsely grouped matrix multiplication. Consider a dense matrix multiplication of matrices $\mathbf{H} \in \mathbb{R}^{a \times b}$ and $\mathbf{W} \in \mathbb{R}^{b \times c}$: 

\begin{equation}
(\mathbf{HW})_{i,j} \delequal \sum_{n=1}^{b}{{h}_{i,n} \cdot w_{n,j}}
\end{equation}
A sparsely grouped version of $\mathbf{W}$ corresponds to a block diagonal matrix $\mathbf{W}^{\it{(G)}}$ with $\it{G}$ groups, a matrix of similar dimension to $\mathbf{W}$  and a sparsity ratio of $1/\it{G}$. This reduces the number of stored parameters, and can be implemented efficiently without zero-multiplication as:

\begin{equation}
(\mathbf{HW}^{\it{(G)}})_{i,j} \delequal \sum_{n=1}^{b}{h}_{i,n} \cdot w_{n,j} = \\
\sum_{n=1}^{b/\it{G}}{{h}_{\;i,\; n \; + \;\tfrac{b}{\it{G}} \; \cdot \; \lfloor j-1\;/\;\tfrac{c}{\it{G}}\rfloor \;} \cdot w_{\;n \;+ \;\tfrac{b}{\it{G}}  \; \cdot \; \lfloor j - 1\;/\;\tfrac{c}{\it{G}}\rfloor ,\;j}}
\end{equation}

An equivalent alternative formulation of a block-diagonal matrix is a grouped convolution for a $1$-dimensional $1\times1$ convolution \citep{squeezebert}. One of our findings is that parameters in the first of the two matricies of the FFN contribute more to task performance, and sparsity is particularly damaging for these fan-out matrices. The second matrix is less sensitive to parameter reduction due to the sparse input and the reduction of projection dimension. Therefore, introducing sparsity in the second matrix results in a Pareto efficient balance between compute and task-performance. The locality constraint of grouped projections on the hidden dimension is detrimental to the model, but this is alleviated by using an output linear projection similar to the output projection matrix used in the MHA block. We find the optimal value for the number of groups to be $G=4$, bringing the parameter count of GFFN to be 75\% of its dense counterpart.

\subsection{Efficient Parameter Utilization}
\begin{figure}[t]
\centering
\begin{subfigure}[t]{0.36\linewidth}
    \includegraphics[width=\linewidth]{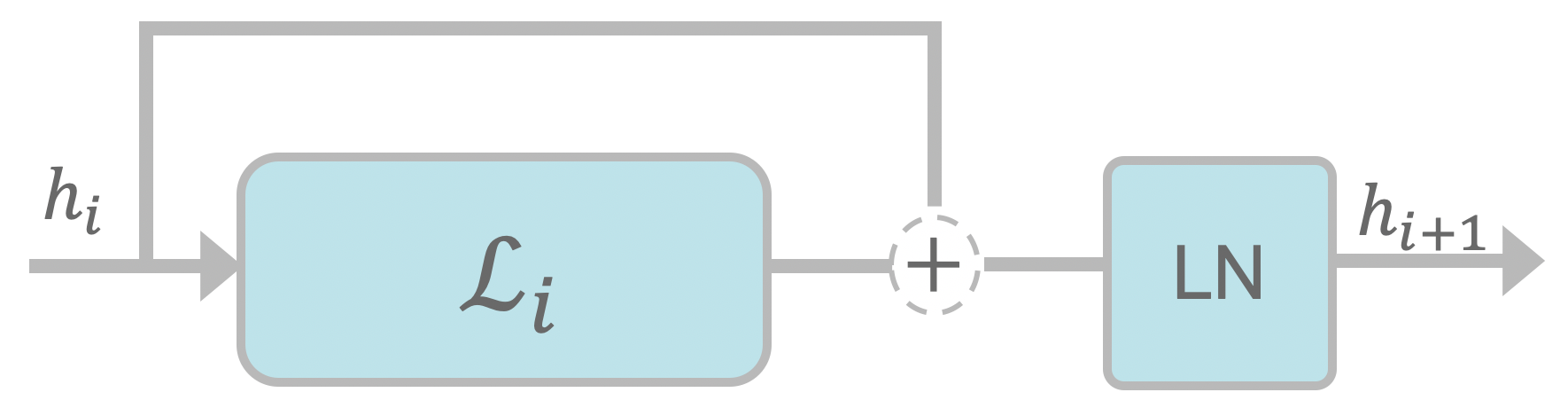}%
    \caption{}
    \label{fig:postnorm}%
\end{subfigure}
\quad
\begin{subfigure}[t]{0.36\linewidth}
    \includegraphics[width=\linewidth]{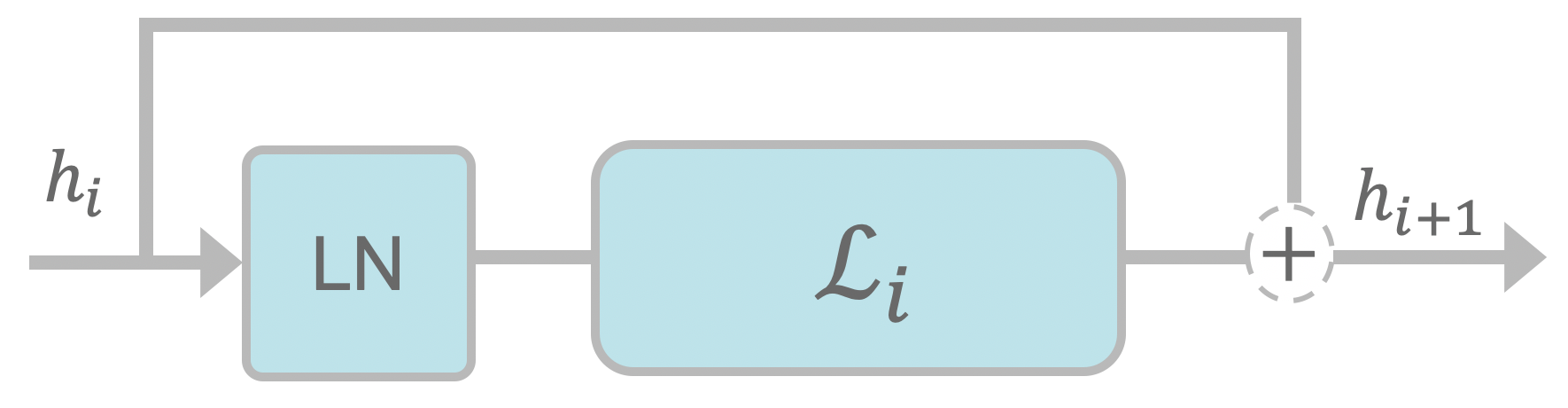}%
    \caption{}
    \label{fig:prenorm}%
\end{subfigure}
\caption{Postnorm (a) and Prenorm (b) module designs for Transformer layer $\mathcal{L}_{i}$.}%
\label{fig:transformer_norms}%
\end{figure}

In line with earlier research on the Transformer architecture \citep{wang2019learning, liu2020understanding, Xiong20}, we move layer normalization \citep{layernorm} from its position after the module's residual (\textit{"postnorm"}, Figure~\ref{fig:postnorm}) to the first position within each residual block (\textit{"prenorm"}, Figure~\ref{fig:prenorm}). While this modification does not directly improve task performance, it stabilizes training and allows the use of a larger learning rate that would otherwise trigger the model with postnorm to diverge. We increase the learning rate by a factor of $4\times$ compared to the postnorm baseline. 

Similarly to \citet{albert}, we find the use of dropout to be detrimental to the pre-training stage. Due to the substantial size of the dataset, this kind of regularization is not required. While removing dropout yields improvements to the pre-training loss, this does not apply to downstream tasks that rely on smaller datasets. Consequently, we include dropout only when fine-tuning on supervised tasks, that have smaller datasets than the pre-training corpus.

\section{Results}

To evaluate the architecture modifications, we chose BERT \citep{bert} pre-training and fine-tuning. The large dataset and challenging training objective mean that task performance improves consistently with model size \citep{albert} and the risk of over-fitting is reduced. This makes it possible to clearly distinguish architecture modifications that benefit efficiency.

Our evaluation of GroupBERT for language representation learning shows that the architecture is:

\begin{enumerate}
    \item Training FLOP-efficient across a range of model sizes (Sections~\ref{sec:pretraining},~\ref{sec:fine-tuning}).
    \item Training time-efficient across a range of compute budgets (Sections~\ref{sec:pretraining},~\ref{sec:fine-tuning}).
    \item Improved by each constituent part (Section~\ref{sec:ablation}).
\end{enumerate}

\subsection{Experiments}
Each experiment consists of two pre-training phases and a fine-tuning phase consisting of multiple training runs, started from the pre-trained model. All phases use the AdamW optimiser \citep{adamw}, with $\beta_1=0.9$, $\beta_2=0.999$, $\epsilon=10^{-6}$. The learning rate follows a linear warm-up decay schedule, whereby the warmup phase lasts for $\min(10^4, 0.1\!\cdot\text{total steps})$ steps, and the peak learning rate depends on the training phase and model size. The model is defined over a vocabulary of $\num{30522}$ WordPiece tokens \citep{wordpiece}. Weights are initialized using a truncated normal distribution of standard deviation $0.02$. For all experiments we use 2 Graphcore M2000 IPU systems.

\emph{Pre-training phase one} optimises the Masked Language Model (MLM) and Next-Sentence Prediction (NSP) loss for corrupted sentence pairs. Masked and padded sequences of length $128$ are grouped into batches of approximately $512$ sequences, with slight variations depending on the model size (see Appendix \ref{app:execution}). The model is trained for $10$ epochs of Wikipedia + BookCorpus \citep{wikipedia, bookcorpus}, corresponding to approximately $8 \!\cdot\! 10^5$ optimisation steps. For all experiments with GroupBERT and the baseline BERT models, the learning rate is set to the largest value that maintains stable convergence. \emph{Pre-training phase two} uses sequence length $384$, $5$ epochs, and approximately $2 \!\cdot\! 10^5$ optimisation steps.

\emph{SQuAD 1.1 fine-tuning} \citep{squad} adds a token span prediction layer and the whole model is fine-tuned to perform extractive question answering. Training uses target batch size $32$ and we train for 2-3 epochs with various learning rates (Appendix \ref{app:hyperparameters}) and report results for the best hyperparameters setting. We report F1 and Exact match scores, which show higher variance than MLM loss values. On the grounds of larger variance, we fine-tune each pre-training checkpoint five times using different seeds for every hyperparameter setting. Fine-tuning has been shown to be quite a brittle process in recent studies \citep{dodge2020fine,zhang2021revisiting,mosbach2021on}. In particular, many instabilities are caused by fine-tuning without using bias correction, an implementation  that was adopted following the original experimental setup of BERT. This omission in the optimizer was observed to cause a collapse of the training process. For this reason, we included a bias-correction term to the AdamW implementation for fine-tuning. 

\subsection{Implementation} \label{sec:implementation}
We train all models on Graphcore Mk2 IPU clusters using a combination of pipeline model parallelism and data parallelism \citep{pipedream}. To make efficient use of compute resources, we increase cluster size and pipeline depth based on model size (see Appendix \ref{app:execution}). We maximise the ``compute'' batch size executed by each node to achieve maximum throughput, and choose the pipeline gradient accumulation count to achieve a target ``global'' batch size.

All parameters, activations, and optimiser states are stored and processed in IEEE 754 half-precision floating-point. This excludes  the Adam weight update calculation and variance state, the loss calculation from logits and temporary variables that require higher precision. Higher precision temporary variables are used by layer normalization, softmax, and the accumulation of partial matrix multiplications. Our models are implemented in TensorFlow and the code will be made publicly available. 

\subsection{Pre-training} \label{sec:pretraining}
BERT type models are notoriously difficult to evaluate as most of the computation is devoted to performing self-supervised pre-training on unlabeled corpora of text. As this process is the most costly one, we are primarily interested in evaluating solely this part of the model training separately. Taking example from other studies \citep{kaplan}, we choose MLM loss on the validation dataset as the most salient criterion for task-performance as it captures language in its generality. 

\begin{figure}[b]
\centering
\begin{subfigure}[t]{0.45\linewidth}
    \includegraphics[width=\linewidth]{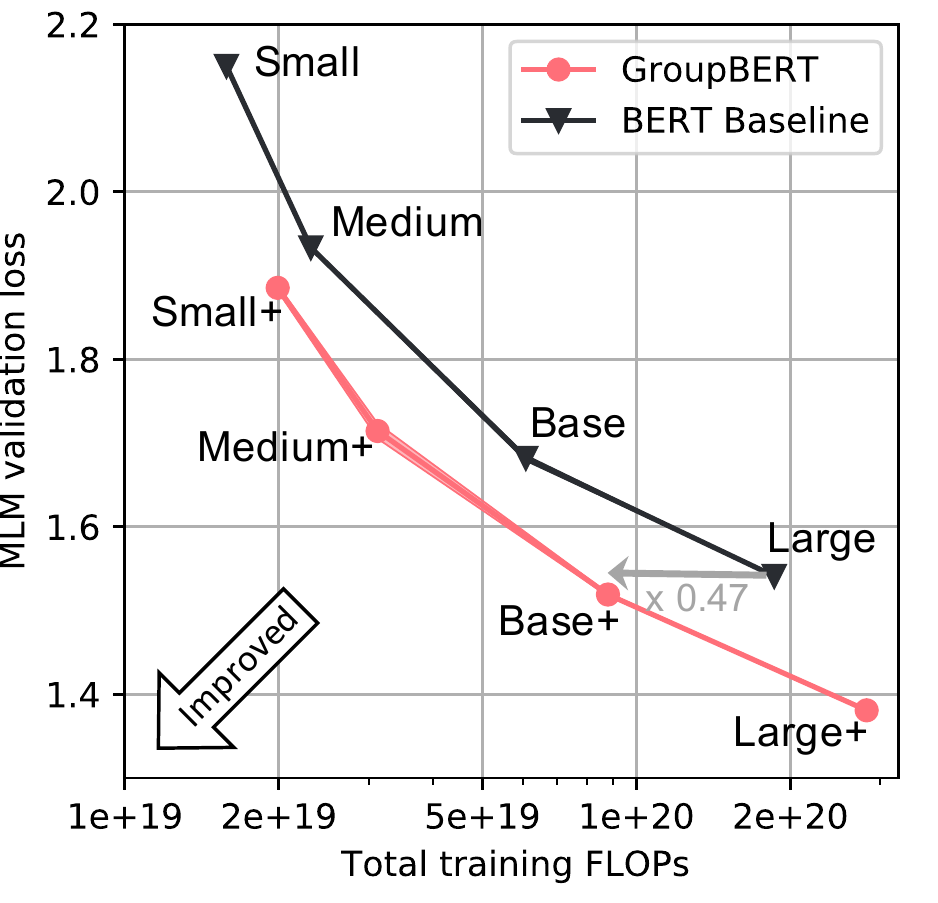}%
    \caption{}
    \label{fig:MLM_Pareto_FLOPs}%
\end{subfigure}
\quad
\begin{subfigure}[t]{0.45\linewidth}
    \includegraphics[width=\linewidth]{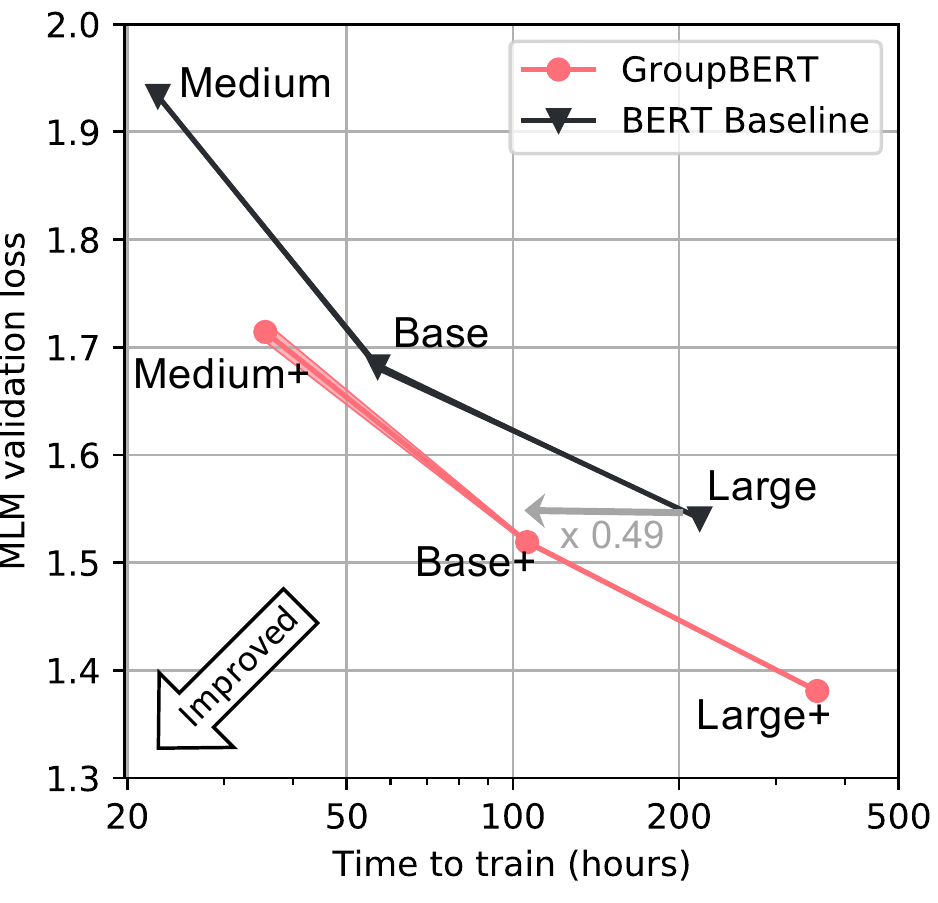}%
    \caption{}
    \label{fig:MLM_Pareto_TTT}%
\end{subfigure}
\caption{The model architecture is Pareto efficient. (a) Mean MLM validation loss $\pm$ standard deviation of the GroupBERT and BERT model families plotted against the total pre-training FLOPs. (b) Mean MLM validation loss $\pm$ standard deviation of the GroupBERT and BERT model families plotted against the total time for pre-training.}
\label{fig:MLM_Pareto}%
\end{figure}

In Figure~\ref{fig:MLM_Pareto} we report the MLM loss measured on the validation dataset for GroupBERT  and baseline BERT models of different sizes.  While evaluating task performance in the context of FLOPs (Figure \ref{fig:MLM_Pareto_FLOPs}) assesses the theoretical performance, the model execution depends on other factors besides FLOPs. For this reason, to complement the performance based on FLOPs, we also assess speed of execution, highlighting that GroupBERT saves overall time-to-train (Figure~\ref{fig:MLM_Pareto_TTT}). Figure~\ref{fig:MLM_Pareto} summarizes the practical benefits of using GroupBERT in terms of both FLOPs and training time to reach a desired task performance. By plotting many model sizes of the same family, the efficiency of GroupBERT is clearly visible. The GroupBERT Base model outperforms baseline BERT Large, but uses only 47\% of its resources and trains in half the time. 

\subsection{Fine-tuning} \label{sec:fine-tuning}
\begin{figure}[htb]
\centering
\begin{subfigure}[t]{0.442\linewidth}
    \includegraphics[width=\linewidth]{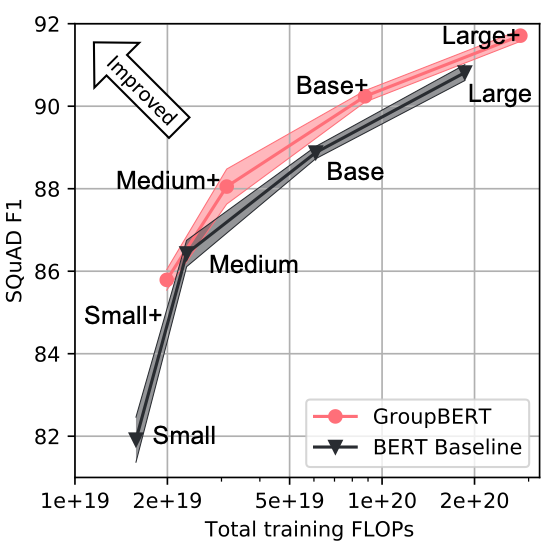}%
    \caption{}
    \label{fig:SQuAD_F1_pareto_FLOPs}%
\end{subfigure}
\quad
\begin{subfigure}[t]{0.45\linewidth}
    \includegraphics[width=\linewidth]{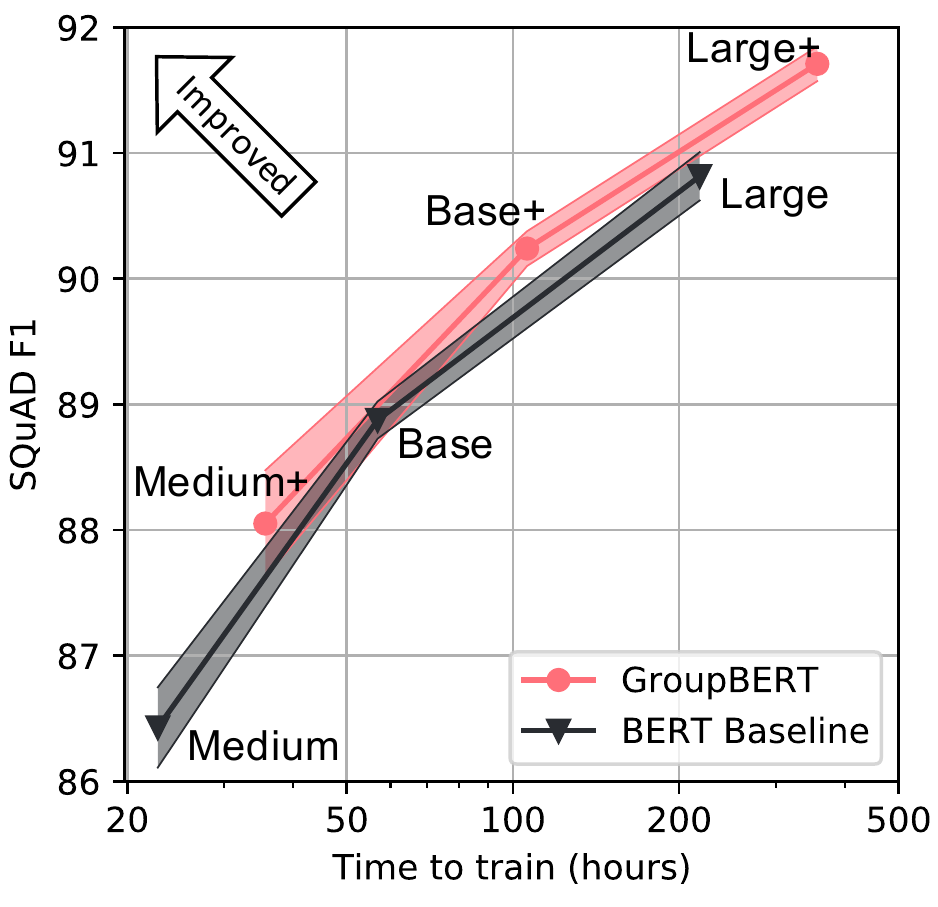}%
    \caption{}
    \label{fig:SQuAD_F1_pareto_TTT}%
\end{subfigure}
\vskip\baselineskip
\centering
\begin{subfigure}[t]{0.442\linewidth}
    \includegraphics[width=\linewidth]{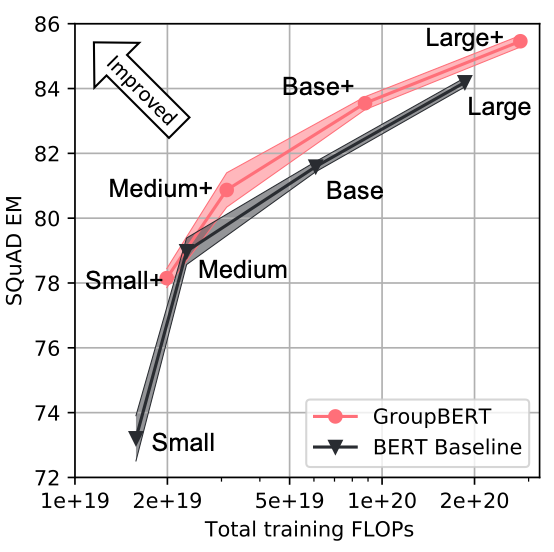}%
    \caption{}
    \label{fig:SQuAD_EM_pareto_FLOPs}%
\end{subfigure}
\quad
\begin{subfigure}[t]{0.45\linewidth}
    \includegraphics[width=\linewidth]{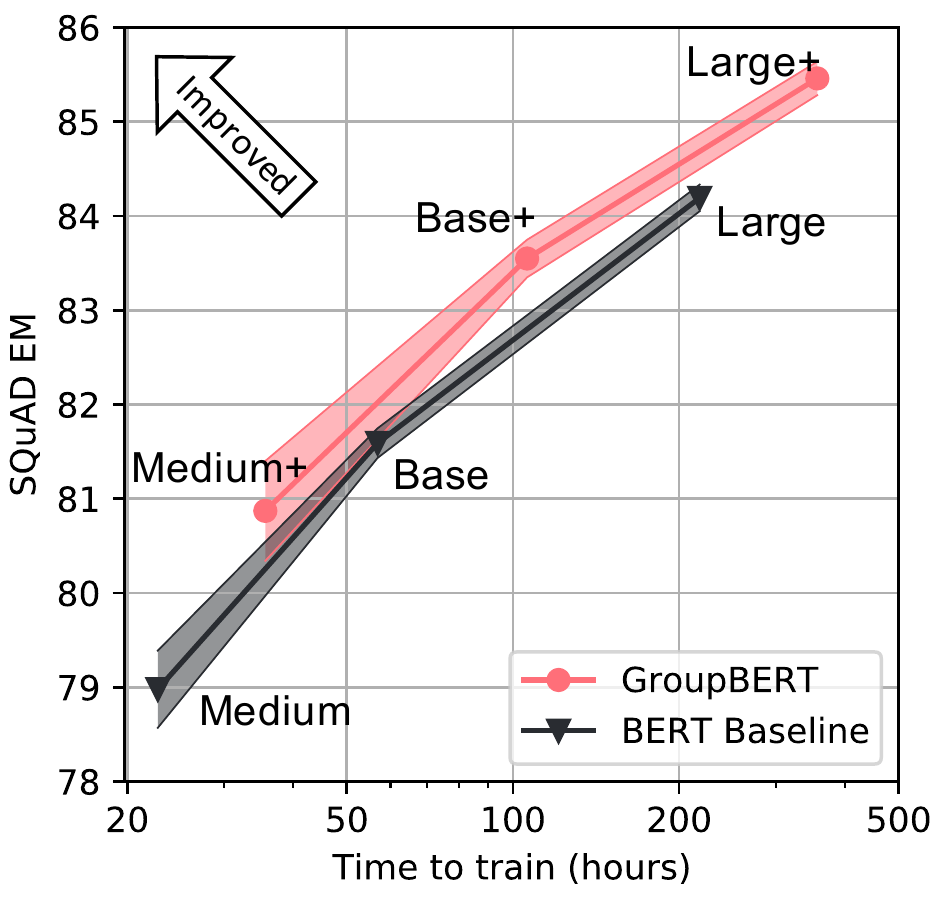}%
    \caption{}
    \label{fig:SQuAD_EM_pareto_TTT}%
\end{subfigure}
\caption{Median SQuAD F1 and Exact Match scores $\pm$ standard deviation of the GroupBERT and BERT model families plotted against the total pre-training FLOPs (a,c) and against the total time to pre-train (b,d).}
\label{fig:SQuAD_pareto}%
\end{figure}

Both the theoretical and the practical Pareto improvements achieved by GroupBERT on MLM evaluation loss translate to the SQuAD fine-tuning task (Figures \ref{fig:SQuAD_pareto} and Appendix \ref{app:hyperparameters}). Not considering the additional dropout in fine-tuning, the cost for training on downstream tasks for different model sizes is proportional to the cost for pre-training, while a few orders of magnitude smaller. Hence, FLOPs and time-to-train as measures of training cost are given for pre-training, as fine-tuning takes a comparatively negligible amount of resources. 

\subsection{Ablation Study} \label{sec:ablation}
To verify the efficiency of the different model changes, we study their individual contribution to the validation MLM loss improvement as well as their cost in terms of parameters and FLOPs (Table~\ref{tab:ModelAblation}). 

Some of the ablation modifications that we make add extra parameters that trigger a divergence during model training when using the best learning rate. Therefore for the ablation study we use the BERT Base model with the next-best learning rate as the main comparison point, to allow the use of the same learning rate throughout. 

\begin{figure} [htb]
\centering
\includegraphics[width=\linewidth]{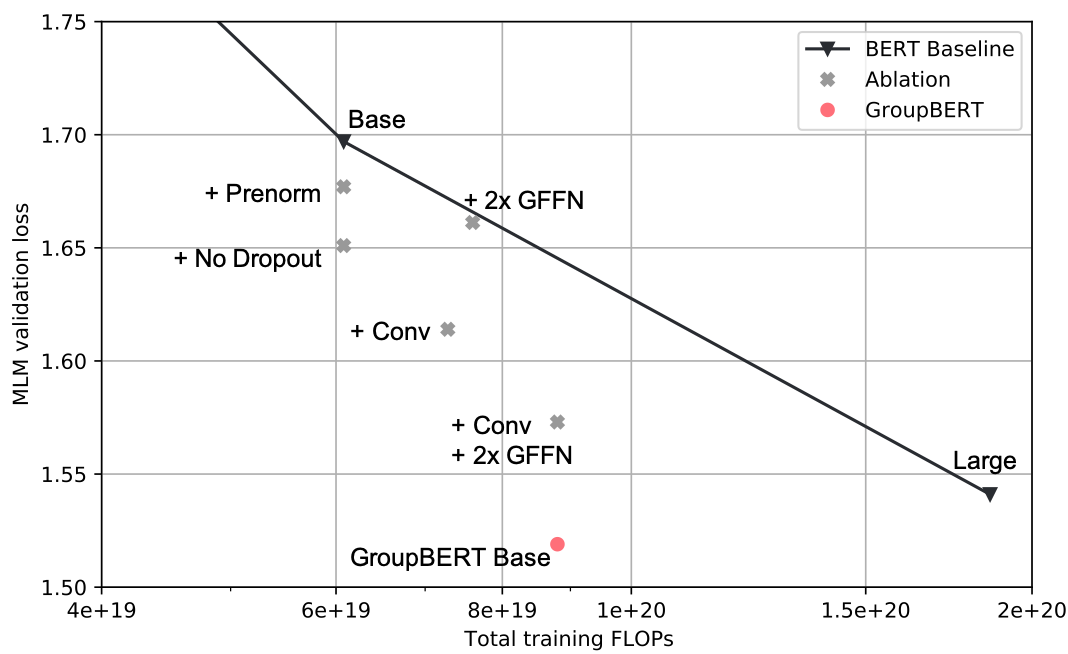}
\label{fig:thtpt1}%
\caption{Ablation results showing the contribution of individual model modifications. Points where we changed the method of training have the same number of FLOPs as the original BERT Base model. Other improvement add some extra FLOPs. }
\label{fig:ablation}
\end{figure}

We measure the performance of each modification with the Pareto improvement of the corresponding model, defined here as the difference in MLM loss with respect to the baseline models for the same FLOPs count. This is visually represented in Figure \ref{fig:ablation} by the vertical distance between the ablation point and the baseline interpolation line.

\begin{table}[htb]
    \caption{Ablation study of the contribution of different model modifications. \vspace{2mm}}
    \centering
    \renewcommand*{\arraystretch}{1.10}
    \begin{tabular}{ l c c c c c} 
        \toprule
        \small \sc Model & \small \sc Parameters & \small \sc Training FLOPs & \small \sc LR & \small \sc MLM loss & \small \sc Improvement \\
        \toprule
        \small BERT Base & \small 110.1M & \small 6.1e19 & \small 2e-4 & \small 1.679 & \small 0.018 \\
        \midrule
        \small BERT Base & \small 110.1M & \small 6.1e19 & \small 1e-4 & \small 1.697 & \small  \\
        \small \quad Prenorm & \small 110.1M & \small 6.1e19 & \small 8e-4 & \small 1.677 & \small 0.020 \\
        \small \quad No Dropout & \small 110.1M & \small 6.1e19 & \small 1e-4 & \small 1.651 & \small 0.046 \\
        \small \quad Convolution & \small 132.4M & \small 7.3e19 & \small 1e-4 & \small 1.614 & \small 0.058 \\
        \small \quad 2 GFFNs  & \small 138.5M & \small 7.6e19 & \small 1e-4 & \small 1.670 & \small 0.005 \\
        \small \quad 2 GFFNs + Conv  & \small 160.8M & \small 8.8e19 & \small 1e-4 & \small 1.573 & \small 0.072 \\
        \midrule
        \small GroupBERT Base & \small 160.8M & \small 8.8e19 & \small 8e-4 & \small 1.516 & \small 0.126 \\
        \bottomrule
    \end{tabular}
    \label{tab:ModelAblation}%
\end{table}

\newpage
 By carrying out individual additive ablation, we are able to see the effect of every model modification in isolation. Furthermore, we investigate the interaction of adding both the convolution module and two GFFN modules. We observe that every module that mixes token information gets significantly more efficient when used with a matching GFFN module. While using two GFFN modules increases the performance in line with the baseline model, using it in conjunction with convolution increases its Pareto improvement.

\section{Discussion}
The improved practical efficiency of GroupBERT reduces the cost of model training. This provides access to better performing language models for a broader range of users, both in the research community and for industrial applications. Moreover, the advantage of GroupBERT over the BERT baseline model family increases with model size, and therefore reduces the cost for training large and highly accurate language models. This observed scaling behaviour indicates the capability of GroupBERT to maintain a material advantage for even larger language models. This can be a valuable improvement for state-of-the-art research, relying on models that are increasing in size exponentially. With the latest breakthroughs using significant amounts of energy and hardware resources \citep{gpt3}, efficiency improvements can translate into significant savings.

The BERT models represent not only a step forward for the NLP research community, but also a tool heavily exploited in day-to-day operations of large corporate entities. Hence, improving the efficiency of these models that are used in production environments can potentially result in significant power savings worldwide, thus limiting greenhouse gas release into the atmosphere. With this goal, all experiments carried out for our investigation were done using $100\%$ renewable electricity, making this work carbon neutral.

In this paper we propose a general technique for improving self-supervised Transformers. However, we only investigate its effectiveness on the English language, due the ease of comparison with other studies using the same datasets. A potential  ethical concern is the increasing dominance of language models trained on the English language, which may be detrimental to the efforts of preservation of endangered languages. Another limitation of this study is the use of the BookCorpus dataset as part of the pre-training process. Pre-training models on this dataset was a standard practice during the course of the investigation, but since then was found to be of questionable ethical value~\citep{addressing_book_corpus}.

We have relied on structured weight sparsity to increase the efficiency of BERT and have created more diversity of building blocks within the self-supervised encoder by complementing the attention block with a convolution module. Our aim is to further build upon these ideas, and construct models that are even more general in their application to different domains and that can be efficiently executed. To do so, we see potential to rely on other forms of sparsity, including dynamic weight sparsity \citep{rigl} and conditional activation sparsity, improving the capability to handle multiple languages and data domains within the same architecture~\citep{switch}.

\section{Conclusion}
In this study, we present GroupBERT: an enhanced Transformer architecture that is shown to be a more efficient alternative, with up to $2.1\times$ efficiency gain in terms of both FLOPs and time-to-train. We achieve these improvements by adding a dedicated grouped convolution module to every layer and using grouped transformations to reduce the density of fully connected layers. The proposed model achieves better results on both pre-training and fine-tuning tasks, and applies to a wide range of model scales.

\newpage
\bibliography{neurips_2021}
\bibliographystyle{neurips_2021}

\newpage

\appendix

\section{Execution scheme} \label{app:execution}

In pipeline parallelism, a model is divided between multiple accelerators. Pipeline parallelism is a form of model parallelism where execution is decomposed into a sequence of stages. At a given instant, every stage is processing a different and independent partition of the model, before passing it on to the next stage. For the majority of our experiments we split the model between four chips, three for encoder layers and one for embedding, projection and loss. However GroupBERT Large has a larger memory footprint, making an eight-chip pipeline the most efficient for execution. To create computational parity between all models, we replicate four-chip pipelines twice for data parallel training, resulting in eight chips being used for all experiments.

In the context of pipelined training, we distinguish between compute batch size and global batch size. The general formula for global batch size is:
\begin{equation*}
    \text{Global batch size} = \text{Replicas} \times \text{Accumulation factor} \times \text{Pipeline depth} \times \text{Compute batch size}
\end{equation*}
Here the compute batch size is the largest portion of the global batch at every pipeline stage, which is always maximized for efficient  resource utilization. The pipeline depth is given by the total number of stages in the forward and backward passes. Therefore, the accumulation factor is the only variable independent from the model structure and pipeline layout, which we can change to get a desired global batch size. We target the use of a global batch size $500\pm20$ for all experiments. This value is not exact and has to vary slightly for different model sizes, as the pipeline layout dictates the exact global batch size that can be used. However, we have not observed any significant effect of these variations on task-performance.


\begin{table}[htb]
    \caption{Overview of the different model sizes used in this study, with the number of parameters used for the BERT baseline model and our GroupBERT architecture. \vspace{2mm}}
    \centering
    \renewcommand*{\arraystretch}{1.10}
    \begin{tabular}{ l c c c c c c} 
        \toprule
        \small \sc Model size & \small \sc Layers & \small \sc Hidden size & \small \sc Params BERT & \small \sc Params GroupBERT\\
        \toprule
        \small Small & \small 4 & \small 512 & \small 29.1M & \small 37.6M \\
        \small Medium & \small 8 & \small 512 & \small 41.7M & \small 56.9M \\
        \small Base & \small 12 & \small 768 & \small 110.1M & \small 160.8M \\
        \small Large & \small 24 & \small 1024 & \small 336.2M & \small 515.5M \\
        \bottomrule
    \end{tabular}
    \label{tab:ModelSizes}%
    \vskip -0.1in
\end{table}

\begin{table}[htb]
    \caption{Pre-training hyperparameters for both BERT and GroupBERT model families. \vspace{2mm}}
    \centering
    \renewcommand*{\arraystretch}{1.10}
    \begin{tabular}{ l l c c c c} 
        \toprule
        \small \sc Family & \small \sc Hyperparameter & \small \sc Small  & \small  \sc Medium & \small \sc Base & \small \sc Large \\
        \toprule
        \multirow{2}{*}{BERT} & \multicolumn{1}{l}{Learning rate} & \multicolumn{1}{c}{4e-4} & \multicolumn{1}{c}{2e-4} & \multicolumn{1}{c}{2e-4} & \multicolumn{1}{c}{1e-4}\\
                                 & \multicolumn{1}{l}{Batch size} & \multicolumn{1}{c}{512}  & \multicolumn{1}{c}{480}  & \multicolumn{1}{c}{480}  & \multicolumn{1}{c}{512}\\\hline
        \multirow{2}{*}{GroupBERT} & \multicolumn{1}{l}{Learning rate} & \multicolumn{1}{c}{3e-3} & \multicolumn{1}{c}{1.5e-3} & \multicolumn{1}{c}{8e-4} & \multicolumn{1}{c}{4e-4}\\
                         & \multicolumn{1}{l}{Batch size} & \multicolumn{1}{c}{512}  & \multicolumn{1}{c}{480}  & \multicolumn{1}{c}{480}  & \multicolumn{1}{c}{480}\\
        \bottomrule
    \end{tabular}
    \label{tab:main_hyperparams}%
\end{table}

\newpage
\section{Fine-tuning Results and Hyperparameters} \label{app:hyperparameters}

 Table \ref{tab:Squad} contains the SQuAD v1.1 fine-tuning results produced from the validation dataset. When fine-tuning GroupBERT models, we always used batch size $32\pm2$ (for the reasons outlined in Appendix \ref{app:execution}), trained for either $2$ or $3$ epochs, with learning rates being one of the following: \{$\operatorname{1e-4}$, $\operatorname{1.5e-4}$, $\operatorname{2e-4}$, $\operatorname{3e-4}$, $\operatorname{4e-4}$\}. Each model required a sweep to identify the best candidate. The sweep for the baseline BERT models was performed according to range of hyperparamters specified in \citet{bert}.

\begin{table}[htb]
    \caption{SQuAD v1.1 results, F1/Exact match \%. \vspace{2mm}}
    \centering
    \renewcommand*{\arraystretch}{1.10}
    \begin{tabular}{ l c c c c} 
        \toprule
        \small \sc Score & \small \sc Small  & \small  \sc Medium & \small \sc Base & \small \sc Large \\
        \toprule
        \small BERT & \small 73.2 / 81.9 & \small 79.0 / 86.4 & \small 81.6 / 88.9 & \small  84.2 / 90.8 \\
        \small GroupBERT & \small 76.8 / 84.7 & \small  80.9 / 88.1 & \small 83.5 / 90.2 & \small 85.5 / 91.7 \\
        \bottomrule
    \end{tabular}
    \label{tab:Squad}%
\end{table}

\section{Attention maps} \label{app:attention_maps}

\begin{figure} [htb]
\centering
\includegraphics[width=\linewidth]{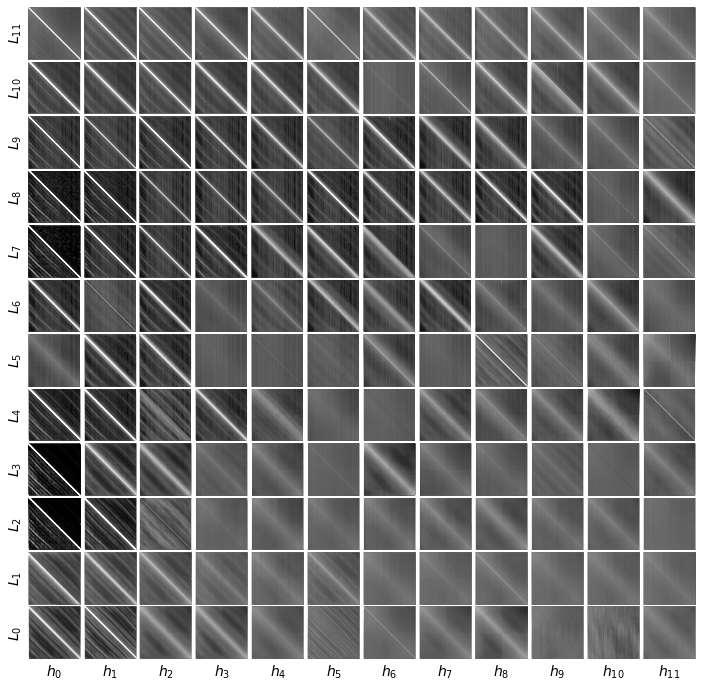}
\caption{BERT Base attention maps, for all layers $L_0 \mathrm{-} L_{11}$. Heads $h_0 \mathrm{-} h_{11}$ in each layer are ordered by positional entropy, Equation~\ref{eqn:headentropy}. Drawn with a value range of $[0, 0.1]$ and gamma of $\nicefrac{1}{3}$.}
\label{fig:all_heads_baseline}
\end{figure}

\begin{figure} [htb]
\centering
\includegraphics[width=\linewidth]{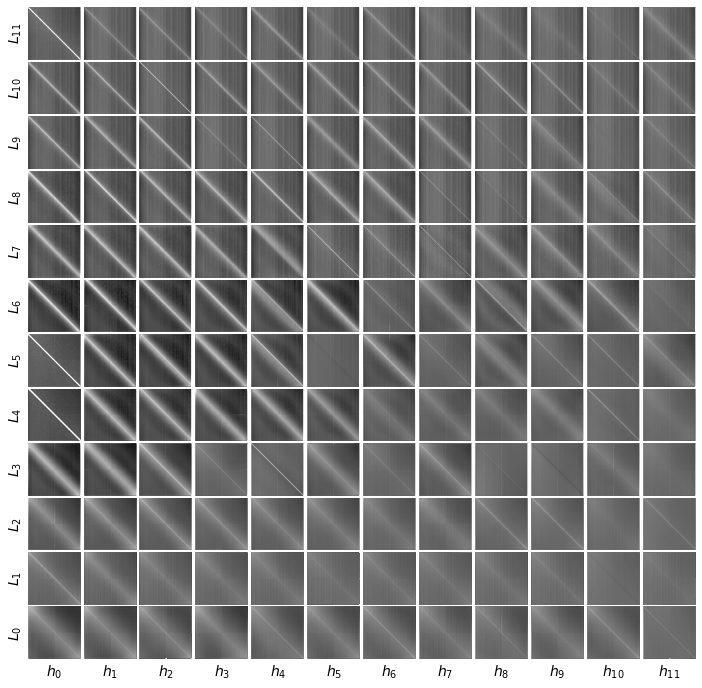}
\caption{BERT Base~+~Conv attention maps, generated and drawn as per Figure~\ref{fig:all_heads_baseline}.}
\label{fig:all_heads_conv}
\end{figure}

\begin{figure} [htb]
\centering
\includegraphics[width=\linewidth]{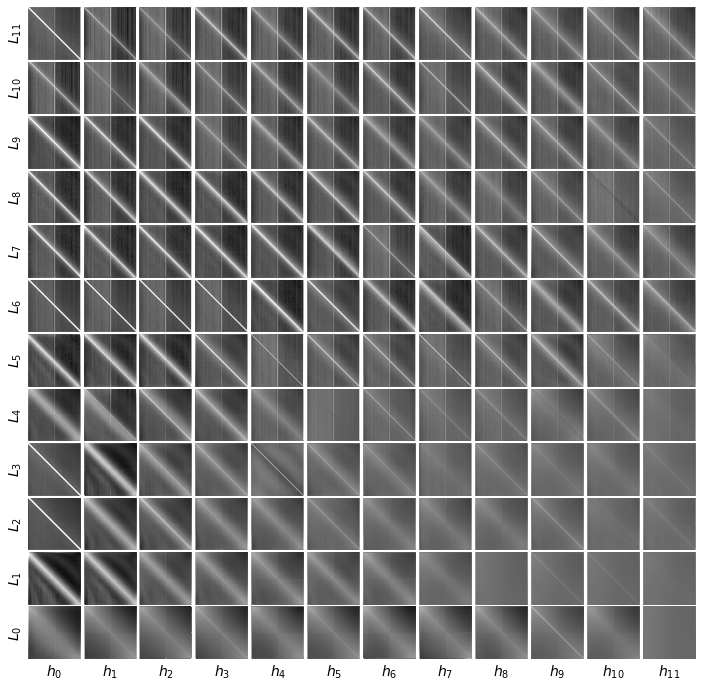}
\caption{GroupBERT Base attention maps, generated and drawn as per Figure~\ref{fig:all_heads_baseline}.}
\label{fig:all_heads_groupbert}
\end{figure}

To visualise the effect of introducing a convolution block into the model, we study the attention maps of each head. An attention map shows the softmax weight between every pair of positions in the sequence. In order to remove the effect of content and focus on position, we average the attention map over $10^3$ validation sequences. We show these maps in Figures~\ref{fig:all_heads_baseline},~\ref{fig:all_heads_conv} and \ref{fig:all_heads_groupbert}, generated after pre-training phase one.

To quantify the locality of an attention head, we define the normalized positional entropy:

\begin{equation} \label{eqn:headentropy}
    H(a) = \frac{1}{L \log L} \sum_{i=1}^L \sum_{j=1}^L a_{ij} \log a_{ij},
\end{equation}

where $a_{ij}$ is the attention weight between source position $i$ and destination $j$ and $L$ is the sequence length. We average the normalized positional entropy over all layers and heads to get a single metric describing a trained model. This measures to what extent attention maps are spread out over different positions. Results for Base-sized models are given in Table~\ref{tab:head_entropy}. The models were trained for the ablation study of Section~\ref{sec:ablation}, showing the effect of individual components on entropy.

\begin{table}[htb]
    \caption{Average normalized positional entropy for the models of Section~\ref{sec:ablation}. Models containing convolutions, Conv and GroupBERT, show higher entropy. \vspace{2mm}}
    \centering
    \renewcommand*{\arraystretch}{1.10}
    \begin{tabular}{ l c } 
        \toprule
        \small \sc Model & \small \sc normalized entropy \\
        \toprule
        \small BERT Base & \small 0.75 \\
        \small Conv & \small 0.92 \\
        \small GroupBERT Base & \small 0.89 \\
        \bottomrule
    \end{tabular}
    \label{tab:head_entropy}
\end{table}

\end{document}